\documentclass[conference]{IEEEtran}
\usepackage{cite}
\usepackage{amsmath,amssymb,amsfonts}
\usepackage{algorithmic}
\usepackage{graphicx}
\usepackage{textcomp}
\usepackage{xcolor}

\usepackage{algorithm}
\usepackage{algorithmic}
\usepackage{subfig}
\usepackage{dsfont}
\usepackage{amsmath}
\usepackage{amsfonts}
\usepackage{newfloat}
\usepackage{listings}
\usepackage{booktabs}
\usepackage{tikz} %
\usepackage[T1]{fontenc}%
\usepackage[utf8]{inputenc}%
\newcommand\copyrighttext{%
  \footnotesize
\textcopyright\ 2025 IEEE. Personal use of this material is permitted.
Permission from IEEE must be obtained for all other uses, in any current or
future media, including reprinting/republishing this material for advertising
or promotional purposes, creating new collective works, for resale or
redistribution to servers or lists, or reuse of any copyrighted component of
this work in other works.} 
\newcommand\copyrightnotice{%
\begin{tikzpicture}[remember picture,overlay]
\node[anchor=south,yshift=10pt] at (current page.south) 
  {\fbox{\parbox{\dimexpr\textwidth-\fboxsep-\fboxrule\relax}{\copyrighttext}}};
\end{tikzpicture}%
}

\def\BibTeX{{\rm B\kern-.05em{\sc i\kern-.025em b}\kern-.08em
    T\kern-.1667em\lower.7ex\hbox{E}\kern-.125emX}}
\begin{document}

\title{Deep Active Re-Labeling: Toward Noise-Resilient Annotation Efficiency}
\author{
\IEEEauthorblockN{Md Abdullah Al Forhad, Weishi Shi}
\IEEEauthorblockA{
\textit{Active Machine Learning Lab} \\
\textit{Learning · Language · Vision (LLaVi) Lab} \\
\textit{Department of Computer Science and Engineering} \\
\textit{University of North Texas, Denton, TX, USA} \\
\{MdAbdullahAl.Forhad, Weishi.Shi\}@unt.edu
}
}

\maketitle
\copyrightnotice

\begin{abstract}
While Deep Active Learning (DAL) effectively reduces human annotation costs, its efficacy is constrained by human annotation errors. This is because the data sampled for active learning is assumed to be highly informative for training. When human annotators introduce errors into this informative data at a certain rate, the active learning performance drops significantly and, in some cases, even exhibits worse outcomes than passive learning. In this paper, we first analyze the impact of human annotation errors in the DAL setting. Then we propose a framework to address the human annotation noise problem for DAL. Informed by human learning patterns, the core idea of our proposed solution involves allocating a portion of the human annotation budget to re-annotate data that has already been labeled. Previous theoretical work suggests that when the model possesses a certain level of ability to identify potentially noisy data, even re-labeling a small fraction of the data can effectively remove noise from the active training set. To achieve this, we implement two active noise sampling strategies to detect noise under different circumstances and allocate a part of the annotation budget to re-annotate these instances. Our approach imbues active learning with a revisiting and introspective behavior. Our experiments demonstrate that, under the same annotation budget, our method is more data-efficient and yields a relatively noise-free annotation dataset in the end.
\end{abstract}

\begin{IEEEkeywords}
Active Re-Labeling, Annotation Efficiency, Deep Active Learning
\end{IEEEkeywords}

\section{Introduction}
\begin{figure}[h]
    \centering
    \subfloat[\centering]{{\includegraphics[width=0.50\columnwidth]{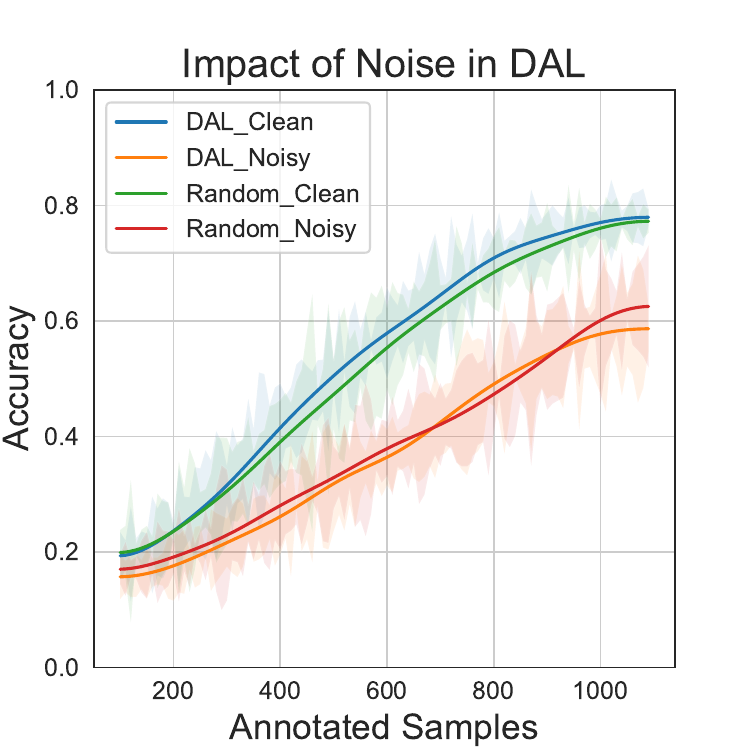} }}%
    \subfloat[\centering]{{\includegraphics[width=0.50\columnwidth]{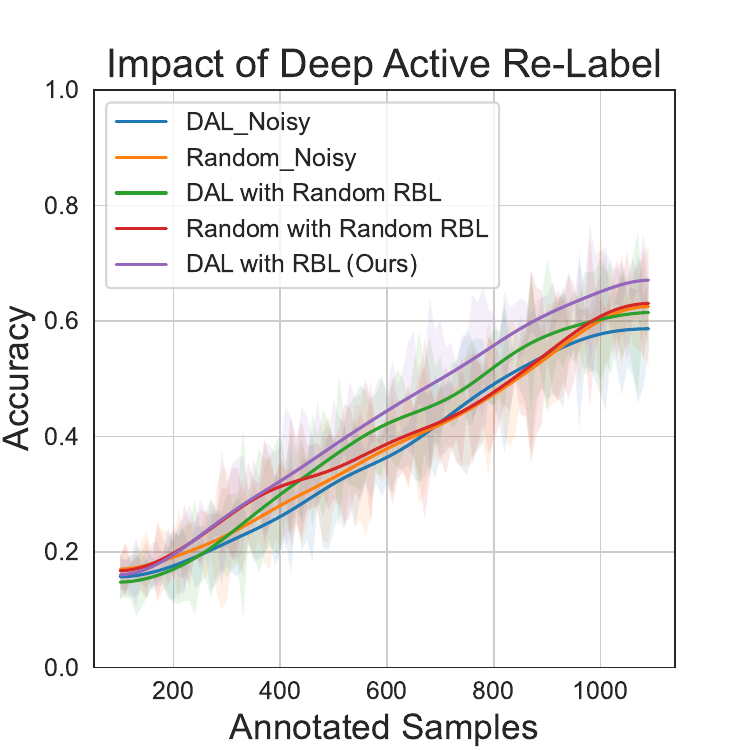} }}%
    \caption{(a) Analyzing the consequences of deep active sampling with noise (DAL\_Noisy) versus without noise (DAL\_Clean), and passive sampling with noise (Random\_Noisy) versus without noise (Random\_Clean) in annotations within DAL. (b) Comparing the effects of re-labeling in deep active learning with active re-labeling (DAL with RBL) and passive re-labeling (DAL with Random RBL), as well as passive learning with passive re-labeling (Random with Random RBL) processes.}%
    \label{fig1}
\end{figure}

Deep learning (DL) models have made it possible for complex tasks to achieve state-of-the-art performance in various fields. However, to be flexible in learning complex patterns, all these models use significantly more parameters than any traditional machine learning model \cite{sanh2019distilbert} \cite{he2016deep}. As a result, the need for massive labeled data is ubiquitous in deep learning, implying a high annotation cost. Active learning (AL) techniques strive to minimize such annotation costs by utilizing a limited number of specifically selected samples for learning \cite{Settles2009ActiveLL}. Several researchers have integrated the concepts of AL and DL and demonstrated that achieving AL objectives is also possible in DL settings. This technique is referred to as Deep Active Learning (DAL) \cite{ren2021survey}. To reduce the annotation cost, DAL method uses various query strategies (\textit{i.e.} Entropy Sampling) to select data to be labeled by human annotators. These query strategies choose the most informative samples to be labeled, which in turn helps the model to achieve generalization with small data. However, there is a strong assumption that the annotator correctly annotates the queried samples. In the real world, this assumption partially holds due to human error, especially when some domain-specific knowledge (\textit{i.e.} Astrobiology and Quantum Computing) is required during the annotation process. 
Figure \ref{fig1}(a) depicts the impact of noisy annotation in the DAL setting. We can observe that a typical DAL strategy outperforms passive learning without annotation noise, as expected. However, its advantage diminishes drastically when moderate noise is present in labels. 
This learning strategy, traditionally considered smart, performs worse under the influence of annotation noise than passive learning, which does not use any strategies at all. This phenomenon of being outsmarted by one's own cleverness is primarily because the DAL model has a tendency to choose the most informative samples for annotation, and these samples have a significant influence on the decision surface. If these crucial samples are incorrectly labeled, they do more harm to the DAL model compared to randomly selected samples.

The issue of noisy annotation in the field of DAL has been overlooked by researchers. Previous work has focused on addressing this issue from different perspectives, such as designing noise-robust model architecture, using multiple oracles, filtering out noisy label annotations, and implementing a dual-purpose learning framework \cite{NEURIPS2021_689041c2} \cite{NIPS2016_299fb214} \cite{younesian2020active} \cite{gupta2019noisy}. However, these approaches have limitations, including the need for budget subsidies, reliance on small hypothetical datasets, and increased complexity due to the use of two distinct models.

Our paper introduces a novel approach that involves integrating relabeling behavior into the DAL process. This behavior is inspired by human introspection, where the brain revisits past learning, identifies inconsistencies, and corrects previous knowledge \cite{karpicke2012retrieval} \cite{clark2013whatever}. To investigate whether the effects of re-labeling annotated labels align with human learning within the DAL, we implement a randomized sample selection method, using the same noisy annotator for the re-labeling process. We assign a small portion of the annotation budget (20 percent) for re-labeling (i.e., total annotation budget remains the same). Figure \ref{fig1}(b) demonstrates that there is very little impact of passive re-labeling on passive learning but the performance improvement of DAL with passive re-labeling is visible. This inspires us to design the re-labeling process systematically. We aim to design a strategy that maximizes the performance of the DAL model while operating within two main constraints: detecting noisy labeled samples and selecting the most informative ones for re-labeling. In this paper, we propose a novel strategy that can help reduce noisy annotation in DAL and help the model achieve generalization with a reduced error bound. The strategy combines the learned representations of the DL model with the spatial properties of a maximum margin classifier (MMC). The proposed strategy (\textit{i.e.} DAL with RBL) converges faster and achieves higher accuracy than any other strategy. Our result suggests that it can effectively address the existing deficiency in DAL with noisy annotations from a unique perspective. Our extensive experiment on real-world datasets provides strong support for our claim. Our contribution is threefold:
\begin{itemize}
\item We systematically analyze the impact of noisy label annotations in the context of deep active learning. Our experimental results validate this analysis. 
\item We introduce the idea of re-labeling the annotated labels with a small portion of the annotation budget in the DAL setting. Our experimental results validate the effectiveness of the proposed idea.
\item We introduce a novel strategy to select samples for re-labeling systematically. This strategy can select wrongly annotated samples with a twofold approach. We show the effectiveness of the proposed strategy with extensive experiments on real-world datasets.
\end{itemize}

\section{Related Work}
Past research in Active Learning has predominantly concentrated on the development of query strategies, with a primary focus on traditional machine learning settings. For instance, \cite{5206627,Tong2001ActiveLT} introduced margin-based uncertainty sampling techniques. However, as deep learning gained prominence, attention shifted to adapting these strategies for deep learning environments. Notably, \cite{gal2017deep} focused on Bayesian neural networks within deep learning settings. Furthermore, several studies have explored effective approaches to train deep learning models with noisy samples across diverse domains, such as \textit{curriculum learning}. A particular area of focus involves selecting accurate samples from noisy data through curriculum learning techniques \cite{jiang2018mentornet} \cite{han2018co}. Some researchers used a meta-learning approach to deal with noisy annotations. Particularly \cite{ren2018learning} \cite{shu2019meta} utilized a meta-learning approach to assign weight to the noisy training set, to balance the impact of noisy samples during the model learning process. \cite{zheng2021meta} used a meta-model called a label correction network to generate corrected labels for noisy data. The main model is then trained using these corrected labels for better learning. \cite{Yu_Shi_Yu_2023} used spatial and temporal properties of the maximum margin classifier to design a resampling strategy in the traditional AL setting. \cite{ducoffe2018adversarial} introduced an adversarial query in the \textbf{Deep Active Learning} domain, with a primary focus on examples lying in proximity to the decision boundary. Although their pseudo-labeling does not corrupt the train set, this approach increases sensitivity in the presence of a noisy annotator. The DAL domain, particularly in the presence of noisy annotation, has received minimal attention. The primary investigation on DAL in the presence of noisy annotations is to build a noise-robust model. \cite{gupta2019noisy} introduced a de-noising layer in their work to build a noise-robust DAL model. A noisy-robust DAL is very challenging because DL models are prone to memorizing label noise, which in turn degrades performance \cite{arpit2017closer}. Another perspective is to use multiple annotators to label the queried samples. \cite{younesian2020active} \cite{Younesian2021QActorAL} proposed to use one weak and one strong annotator. Their initial model was assumed to have been trained on a very small set of noise-free data, which is a strong and unreliable assumption. \cite{goh2023activelab} proposed a method that can select which data to label, allowing for labeling and relabeling, specifically designed for multiple annotators. 
Using multiple annotators is impractical in fields with scarce experts and increases costs. Majority voting, often relied upon, is less effective in specialized domains like Astrobiology or Quantum Computing, where experts vary in knowledge. The assumption of equal contribution among experts is unreliable, and assigning appropriate weights is difficult due to the limited number of experts.
Moreover, most approaches addressing annotation noise with multiple annotators are costly and impractical. Our proposed strategy, however, is more practical, reduces label noise, and leads to faster convergence and higher accuracy in DAL models.

\section{Methodology}
\subsection{Problem Setup}
We focus on a pool-based active learning setting. Let $\mathcal{D}_T = {\{(\textbf{x}_n,y_n)\}}_{n=1}^N$ and $\mathcal{D}_P = \{\textbf{x}_m\}_{m=1}^M$ denote the annotated dataset pool and an unlabeled dataset pool, where $N << M$. Initially, a model $\mathcal{M}$ is trained on $\mathcal{D}_T$. In each DAL round, an acquisition function $f(\textbf{x},\mathcal{M})$ with $\textbf{x} \in \mathcal{D}_P$ selects $\textbf{x}^*$. 
\begin{equation}
\textbf{x}^* = \underset{\textbf{x} \in \mathcal{D}_P}{\arg\max} f(\textbf{x},\mathcal{M})
\end{equation}
Afterward, the new instance-label pair $\{(\textbf{x}^*,h_{\alpha}(\textbf{x}))\}$ is added to $\mathcal{D}_T$ and the model $\mathcal{M}$ re-trained based on the updated $\mathcal{D}_T$. 
We assume $h_\alpha(.)$ is a noisy annotator and $\alpha$ denotes the degree of annotation noise. Denote $\alpha = p(h_\alpha (\textbf{x}) \neq h_o(\textbf{x}))$. Denote $h_o(.)$ as a true annotator that can be achieved as a noise rate approach to zero, $h_o(\textbf{x}) = \lim\limits_{\alpha\to0} h_\alpha(\textbf{x})$. We also assume that the initial $\mathcal{D}_T$ is also noisy at the same rate as $\alpha$. 
\subsection{Re-Labeling for Label Correction}
Throughout the annotation process, different forms of interference may occur as a result of the annotator's restricted viewpoint or perception, difficulties in accurately assessing the labeling function $h_0$, or the existence of uncertainties such as random errors. In a classification problem with $K$ classes, we assume that when annotation noise is present, all the noisy classes have an equal chance of being wrongly assigned to the data. In other words, $p(y=y_k|h_0(\textbf{x}) \neq h_{\alpha}(\textbf{x})) = \frac{1}{K-1}$.
One direct solution to reduce the annotation noise is to lower the level of annotation error rate, $\alpha$. Nevertheless, this approach presents challenges as it requires addressing the error behavior from the human side. Reducing human error, particularly in specialized fields like medical image labeling, can be formidable and resource-intensive when relying solely on conventional methods such as training, domain expertise, and accumulated experience. Additionally, the annotation error rate, $\alpha$, is expected to be lower than uniform errors as the annotator for the DAL setting is a domain expert. This infrequent error occurrence can be viewed as a rare event stemming from human cognitive
bias. Increasing human interaction can help counter some biases \cite{KLIEGR2021103458}. DAL setting utilizes human interaction for labeling, and we aim to increase the interaction for label correction through re-labeling. 
%This leads to an alternative yet highly effective strategy for mitigating labeling errors that involves leveraging the same annotator to re-evaluate samples that have undergone prior annotation \cite{Yu_Shi_Yu_2023}. 
When annotators label and re-label data repeatedly, they are more likely to notice inconsistencies or errors in their own judgments.  This approach is practical in real-world scenarios and fiscally prudent, as it eliminates the need for an additional annotator, making it suitable for both single and multi-expert settings. Among the most straightforward techniques for re-labeling is the random selection of samples from $\mathcal{D}_T$ to reassign labels.
The results depicted in Figure \ref{fig1}(b) show that even this most straightforward approach enhances the model's performance compared to the absence of re-labeling. To find the candidate for resampling, we introduce a re-label acquisition function $q(): \mathcal{X} \rightarrow \mathcal{R^{+}}$ that selects samples from $\mathcal{D}_T$ and presents them to $h_\alpha(.)$.
\begin{figure}[h]
    %\centering
    \subfloat[\centering]{{\includegraphics[width=0.33\columnwidth]{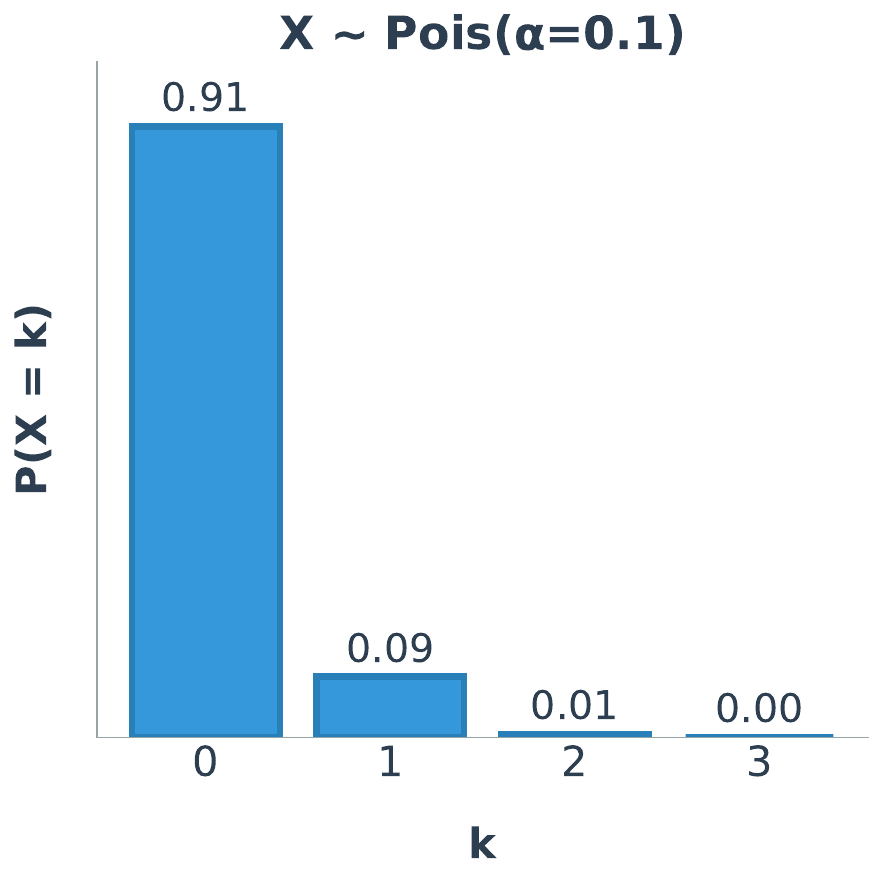} }}%
    \subfloat[\centering]{{\includegraphics[width=0.33\columnwidth]{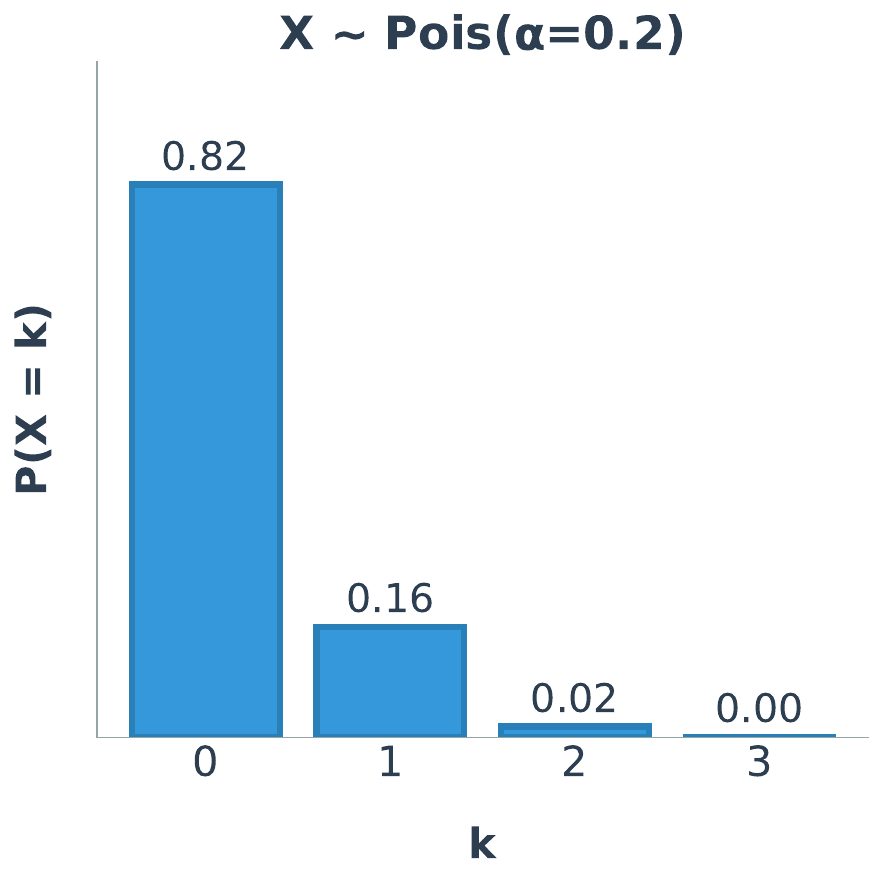} }}%
    %\hfill
    %\subfloat[\centering]
    %{{\includegraphics[width=0.30\columnwidth]{2c.pdf} }}%
    \subfloat[\centering]{{\includegraphics[width=0.33\columnwidth]{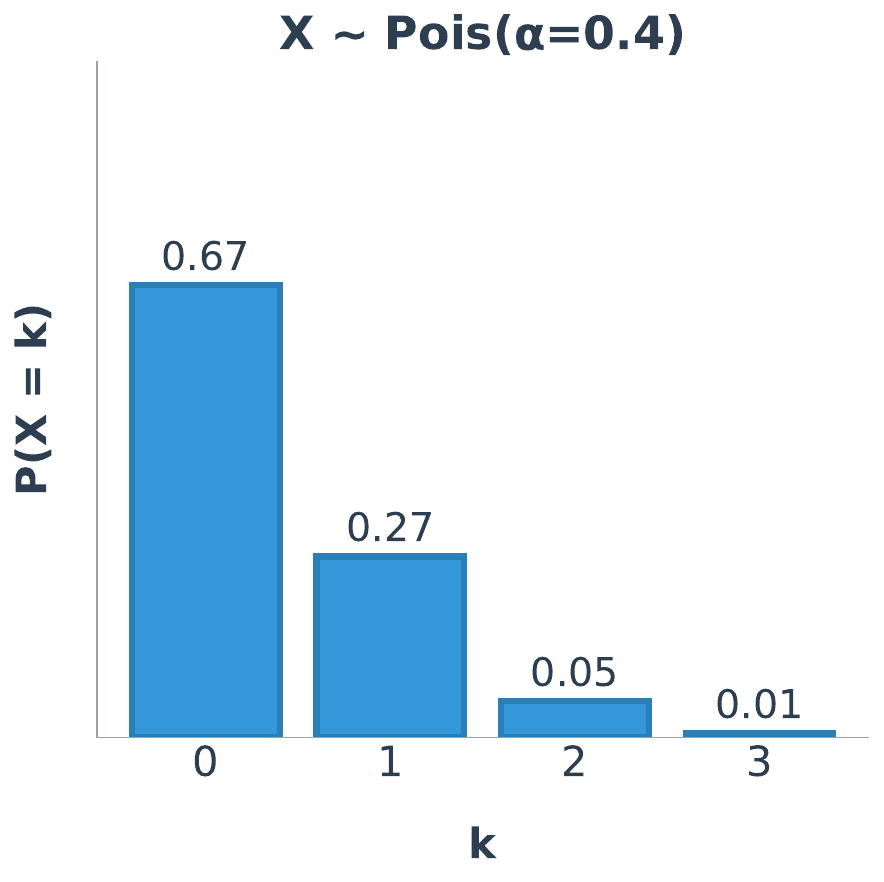} }}%
    \caption{Relationship between re-labeling count and incorrect annotation probability. (a) With $\alpha$ set to 0.1, the probability of encountering an incorrect label during the third re-labeling (k=3) diminishes completely. A similar trend continues for (b) with $\alpha$ set to 0.2. This consistent trend persists even with a relatively higher $\alpha$ value of 0.4, as demonstrated in (c).}%
    \label{fig2}
    \vskip -0.15in
\end{figure}
\begin{figure*}[t]
\centering
\includegraphics[width=0.90\textwidth]{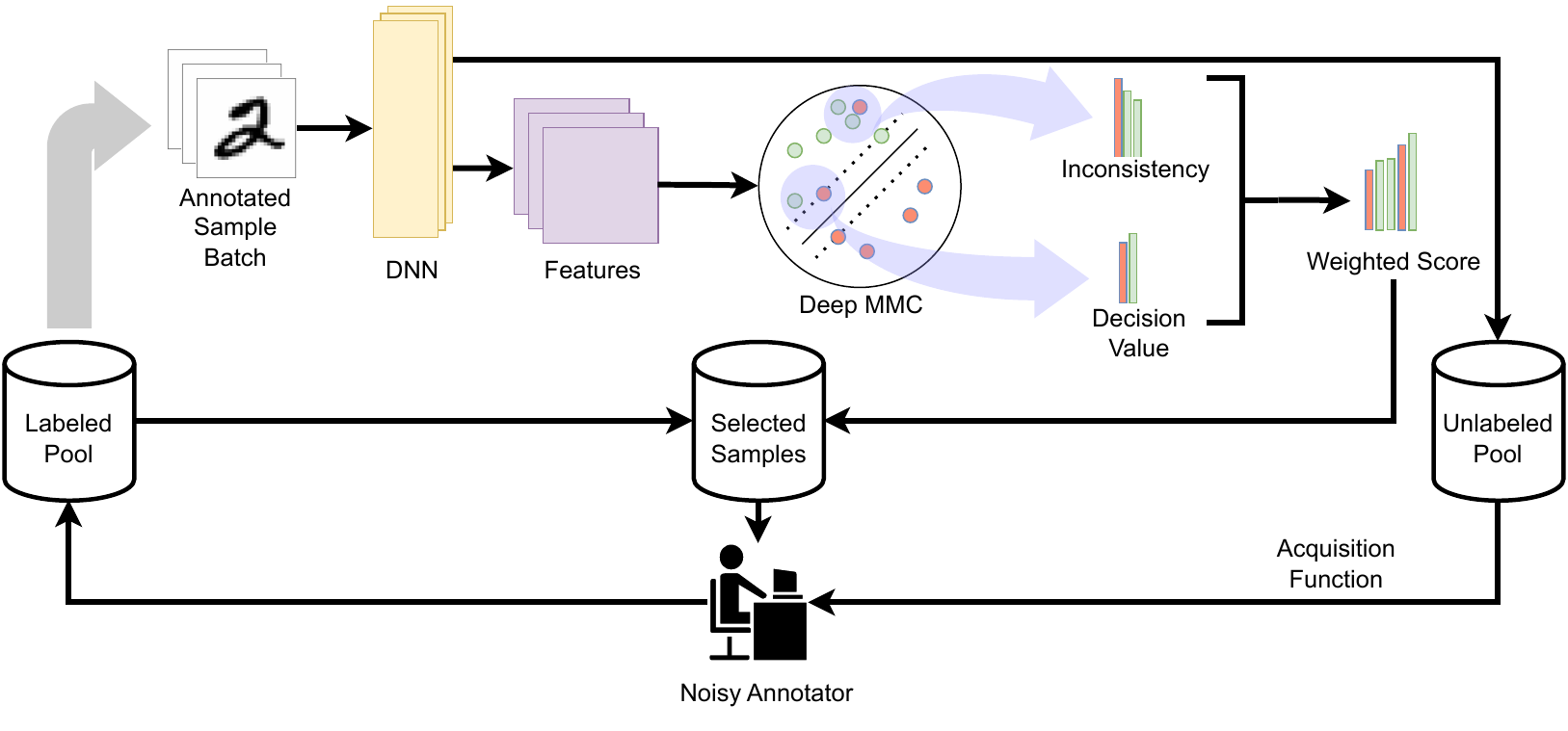}
\caption{Our proposed framework for deep active re-labeling in DAL setting.}
\label{fig3}
\end{figure*}
\section{Active Re-Labeling Strategy}
In this section, we first present a mathematical analysis based on the Poisson distribution to establish the theoretical foundation, then develop a systematic strategy.
\subsection{Mathematical Analysis.} 
Randomly double-checking labeled instances can reduce the noise rate in the training dataset, but a more principled inspection strategy is more effective. Suppose we create a re-label acquisition function $q^*$ that selectively identifies potentially mislabeled samples, where $q^*$ is allowed to re-select the same sample multiple times. If a given instance is re-labeled $k$ times, a natural question arises: \textit{What is the probability that the instance remains incorrectly annotated after these repeated inspections?} 
We model this using the probability mass function of the Poisson distribution, 
$
P(X=k) = \frac{e^{-\alpha} \alpha^k}{k!}, 
$
where $e$ denotes Euler’s constant, $\alpha$ is the noise rate of the annotator, and $P(X=k)$ represents the likelihood of observing $k$ incorrect annotations for a single sample. The Poisson distribution is a principled choice here because annotation errors, especially by domain experts, are typically rare and independent events—conditions under which Poisson processes are commonly employed in fields such as biology, communication theory, and reliability engineering. 
The analytical advantage of this framework is that it provides a clean, \textbf{closed-form} characterization of how the probability of persistent mislabeling decays as the number of re-labeling attempts increases. As illustrated in Figure~\ref{fig2}, the probability of a sample being incorrectly labeled decreases sharply with every re-labeling, and when $k > 2$, the probability already falls below the initial error rate $\alpha$. This trend persists even at higher values of $\alpha$, showing that repeated re-labeling guarantees a reduction in the effective noise rate. 
While real-world annotator errors may also include non-uniform or correlated biases, the Poisson model should be viewed as a theoretically tractable first-order approximation that captures the essential mechanism of error reduction through repeated inspection. In practice, our empirical results (see Sec.~\ref{flipt}) confirm that the qualitative behavior predicted by this model holds even under more complex noise scenarios, thereby validating its use as a theoretical foundation for our framework. 

This leads to another crucial question: \textit{Should we rely on the same annotator multiple times, or should we distribute re-labeling across multiple annotators?} To answer this, consider a multi-annotator setting with $N$ experts, each characterized by a noise rate $\alpha_i$. Let $p(\alpha_i)$ denote the probability of selecting annotator $i$, where $\sum_{i=1}^{N} p(\alpha_i) = 1$. In practice, annotators with higher accuracy are more costly, so $p(\alpha_i)$ typically decreases as $\alpha_i$ decreases. The expected noise rate of the system can then be expressed as
\(
\alpha = \sum_{i=1}^{N} p(\alpha_i) \cdot \alpha_i.
\)

Now, consider the special case where a single annotator with noise rate $\alpha$ is decomposed into two hypothetical annotators, $\alpha_1$ and $\alpha_2$, with selection probabilities $p(\alpha_1)$ and $p(\alpha_2)$, respectively, such that $p(\alpha_1) + p(\alpha_2) = 1$. In this setting,
\(
\alpha = p(\alpha_1) \cdot \alpha_1 + p(\alpha_2) \cdot \alpha_2,
\)
which is mathematically equivalent to repeatedly querying the same annotator with noise rate $\alpha$. 

This observation provides a useful justification: \emph{re-labeling by the same expert can be treated as a probabilistic equivalent of querying a pool of multiple annotators}. Importantly, this equivalence makes our strategy practical in budget-constrained environments, where recruiting multiple experts is infeasible. Instead of requiring access to several annotators with varying expertise levels, one can obtain similar error-reduction benefits by strategically re-using the same annotator through a well-designed re-label acquisition function $q^*$. 

Of course, constructing an accurate and stable $q^*$ is non-trivial. Its behavior resembles a complex, high-dimensional black-box function, since annotation errors are influenced by annotator bias, data difficulty, and noise correlations. To address this, we propose leveraging deep learning structures to guide the design of $q^*$, enabling systematic identification of noisy labels rather than relying on random re-labeling or impractical multi-annotator assumptions.

\subsection{Systematic Strategy Design}
The thorough analysis and reliable mathematical assurances have inspired us to develop a systematic approach to re-labeling in DAL in the presence of a noisy annotator. To design such a strategy as stated above we can detect noisy samples based on the probability of being most likely to least likely noisy samples. To detect noisy labels there has been extensive research in the research community. On many occasions, Maximum margin classifiers (MMC) have been proven very successful in detecting label noise \cite{biggio2011support,6460920}. The previous work was based on the properties of MMC~\cite{Yu_Shi_Yu_2023}. The work yielded theoretical results with MMC, which accomplished guaranteed error reduction but was solely applicable in traditional machine learning settings. This limitation arises from the classifiers' constraints when faced with an expanding number of feature spaces. To leverage the effectiveness of MMC within the DAL setting, we utilize the learned feature representations of samples from the DAL model and transfer them to MMC. By doing so, we extend the capabilities of the DAL model to the traditional MMC. The entire proposed framework is depicted in Figure \ref{fig3}. Here we can see the traditional DAL process with additional steps that involve our novel approach to introduce the re-labeling process. 

\subsubsection{Connecting MMC and DNN} 
We aim to detect noisy labels from the deep neural network (DNN) while utilizing the geometric properties of the MMC. To establish a connection between DNN and MMC, we first provide a brief overview of the relationship between neural networks and Gaussian processes (GP) \cite{Neal1995BayesianLF} \cite{Neal1996PriorsFI} \cite{10.5555/2998981.2999023} \cite{lee2018deep} \cite{khan2020approximate}. Let's assume we have a fully connected neural network with $L$ hidden layers with width $N_l$. For a single-hidden layer neural network the ith component of the network output, $z_i^1$, is computed as:
\begin{equation}
%\begin{split}
\small
z_i^1(x)=b_i^1+\sum_{j=1}^{N_1} W_{i j}^1 x_j^1(x),  \quad x_j^1(x)=\phi\left(b_j^0+\sum_{k=1}^{d_{i n}} W_{j k}^0 x_k\right),
%\end{split}
\end{equation}

Here, $\phi$ denotes non-linear activation and $d_{in}$ is the input dimension ($x \in \mathbb{R}^{d_{\text{in}}}$). The weight $W_{ij}^1$ and the bias $b_i^1$ of the neural network are i.i.d. parameters.  Due to the i.i.d. assumption, the post-activation values (denoted $x_j^1$ and $x_{j^\prime}^1$)  for different neurons $j$ and $j^\prime$ are independent. The Central Limit Theorem states that the sum of a large number of independent random variables converges to a Gaussian distribution. So the pre-activation values $z_i^1(x)$ of neurons in the first layer, when summed over all inputs, converge a Gaussian distribution as the width of the layer $N_1$ approach to infinity. Because the pre-activations approach a Gaussian distribution, finite collections of these pre-activations for different inputs jointly form a multivariate Gaussian distribution. More precisely, the activation as a function defined in the continuous input domain should follow a Gaussian process (GP). 
Denote $\mathcal{GP}(\mu, K)$ a Gaussian process with mean function $\mu(.)$ and covariance function $K(. , .)$. So, the preactivation values ($z_i^1$) of neurons in the first layer follow a Gaussian process ($\mathcal{GP}(\mu^1,K^1)$).
Note that, these properties are invariant to neuron index $i$ perturbation and the parameters possess a zero mean, it follows that mean 
%\begin{equation}
$ \mu^1(x)=\mathbb{E}\left[z_i^1(x)\right]=0 $
%\end{equation}
and covariance:
\begin{equation}
K^1\left(x, x^{\prime}\right) \equiv \mathbb{E}\left[z_i^1(x) z_i^1\left(x^{\prime}\right)\right]
\end{equation}
The argument holds as we increase the number of layers $l$ and for each layer $N_l \rightarrow \infty$. Then we will have a joint multivariate Gaussian distribution, $z_i^l \sim \mathcal{GP}(0, K^l)$ where the covariance is: 
\begin{equation}
K^l\left(x, x^{\prime}\right) \equiv \mathbb{E}\left[z_i^l(x) z_i^l\left(x^{\prime}\right)\right]
\end{equation}
When we provide a GP with a covariance function, it enables us to effectively use an infinite number of features. The Radial Basis Function (RBF) is an example of such a kernel, as it possesses an infinite number of features. This implies that we can use the neural network as a feature extractor and plug its outputs into a GP for classification tasks. Specifically, the outputs of the neural network $z_i^l$ can serve as the input features for the Gaussian process.
\begin{equation}
z_i^l(x) \sim \mathcal{GP}(0, \textbf{K}_{RBF}^l\left(x, x^{\prime}\right))
\end{equation}
Similarly, MMCs leverage the kernel trick to compute the inner product of input vectors in a higher-dimensional space without explicitly transforming them into that space. Both MMCs and GPs employ kernel functions to capture the similarity between data points. Remarkably, the RBF kernel, commonly used in Support Vector Machine (SVM), which is a type of MMC, is equivalent to the covariance function of a GP with specific parameters. This suggests that if the features learned by a DNN can be used in GP, then the features can also potentially be employed to model a sparse kernel machine (\textit{i.e.} MMC).

\begin{figure}[t]
    \centering
    \subfloat[\centering]{{\includegraphics[width=0.49\columnwidth]{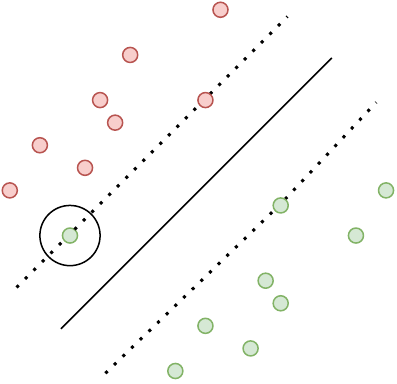} }}%
    \subfloat[\centering]{{\includegraphics[width=0.49\columnwidth]{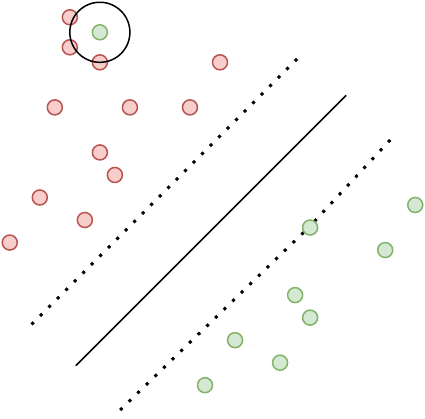} }}%
    \caption{Different types of noisy annotations. (a) A noisy annotated sample with the lowest distance from the current decision surface. (b) A Noisy annotated sample is on the wrong side of the margin and has a different label than its neighboring samples.}%
    \label{dec}
\end{figure}
\subsubsection{Model Formulation.} We proceed with the training of the DAL model denoted as $\mathcal{M}_{\theta_1}$. This involves optimizing the model parameters $\theta_1$ through the minimization of a chosen loss function, typically the cross-entropy loss, employing an optimization strategy, such as Stochastic Gradient Descent (SGD). The detailed structure of the model is discussed in the experiment section. Subsequently, we perform an additional training phase for MMC, denoted as $\mathcal{M}_{\theta_2}$, built upon the refined parameters $\theta_1$. This involves minimizing the MMC's designated maximum-margin loss function, often the hinge loss. Consequently, the formulation can be expressed as follows:
\begin{equation}
g(\textbf{x}) = \mathcal{M}_{\theta_2}(\mathcal{M}_{\theta_1}(\textbf{x})) = {\theta_2}^T \phi(\textbf{x})
\end{equation}
Where ${\theta_2}^T \phi(\textbf{x})$ is a MMC and $\theta_2$ can be learned  using 
\begin{equation}
\arg\min_{\theta_2} \sum [1 - y_n g(\textbf{x}_n)] + \lambda||\theta_2||^2 
\label{eq3}
\end{equation}
Where the first part of the equation (\ref{eq3}) denotes hinge loss. As the MMC utilizes the learned feature space of DAL, we refer to it as the deep MMC. When the deep MMC makes an incorrect prediction or doesn't have enough separation between different classes, the hinge loss function imposes a penalty. The size of the margin determines the degree of penalty - larger margins result in lower hinge loss. So this deep MMC attempts to find a decision boundary that maximizes the margin. When there is label noise, correctly labeled samples are likely to have a larger margin, while incorrectly labeled samples tend to be closer to the boundary. Figure \ref{dec}(a) depicts this type of noise. We can measure the perpendicular distance from the decision surface to detect this type of noise. This gap can be measured using the decision function of deep MMC, we define:
\begin{equation}
q^{DEC}(\textbf{x}) = | \theta_2^T \phi(\textbf{x}) | \propto \frac{| y \theta_2^T \phi(\textbf{x}) |}{||\theta_2||} 
\label{eq4}
\end{equation}
Where DEC stands for decision and the last part of the equation (\ref{eq4}) denotes the perpendicular distance from the decision surface \cite{bishop2006pattern}. We've leveraged the fact that for a binary problem, $y \in \{-1, 1\}$, and the norm of $\theta_2$ remains constant for all input values $\textbf{x}$ when the learning process of deep MMC fully converges. To extend this to a multiclass problem we have utilized with One-vs-the-rest (OvR) multiclass strategy.

The function $q^{DEC}(\textbf{x})$ is useful for identifying margin noise, but other types of noise may exist outside of the margin. To identify this type of noise, we check if the labels of neighboring samples differ from the label of the specific sample. Figure \ref{dec}(b) depicts this type of noise which assesses label disagreement for input \textbf{x} using variable \textbf{y}. We define $q^{INC}(\textbf{x})$ to calculate the score error associated with predicting the output $y$ for a given input $\textbf{x}$, with respect to the MMC's decision boundary $|\theta_2^T \phi(\textbf{x})|$ and $\theta_2$ by utilizing the kernel function $k(\textbf{x}_i,\textbf{x}_j) = \phi(\textbf{x}_i)^T \phi(\textbf{x}_j) $. We propose:
\begin{equation}
q^{INC}(\textbf{x})=\sum_{(\textbf{x}_i)\in\mathcal{D}_T}|| k(\textbf{x},\textbf{x}_i) y - \theta_2^T \phi(\textbf{x})||   
\end{equation}
Where INC stands for inconsistent. This $q^{INC}(\textbf{x})$ function will help detect inconsistent samples within the training set. A large score means the sample is more inconsistent than others. 

To combine both $q^{DEC}(\textbf{x})$ and $q^{INC}(\textbf{x})$ function we use combined weighted score. We first normalize the scores before computing their linear combination. The decision value, although it varies between tasks, is appropriately scaled and reflects the re-labeling importance from the decision perspective. We introduce a dynamic weighting scheme because of the special training process of MMC. The MMC algorithm aims to find the hyperplane that maximizes the margin, and this involves continuously adjusting the margin until it reaches the maximum. During the initial stages of training, the margin tends to fluctuate more as the algorithm strives to identify the most suitable hyperplane. However, as the training advances, the margin becomes more stable as the algorithm converges on a solution. So during the initial DAL process, the label correction of margin noise will further speed up the model convergence. We propose to include a weighted factor, $\tau \in (0,1)$. This factor starts from zero and updates during the DAL process. Initially, $\tau$ will favor $q^{DEC}(\textbf{x})$ and penalize $q^{INC}(\textbf{x})$. With each round of DAL, $\tau$ will gradually shift its weight towards $q^{INC}(\textbf{x})$.
\begin{equation}
q^{RBL}(\textbf{x}) = (1 - \tau) q^{DEC}(\textbf{x}) + \tau [q^{INC}(\textbf{x})]^{(-1)}
\label{eqr}
\end{equation}
Where RBL stands for re-label. This (\ref{eqr}) now concentrates on different types of noise across different rounds of the DAL process. It first identifies wrongly annotated samples on the margin, then gradually shifts its focus to identifying other inconsistent label noise samples.
%As the DAL process advances, the function $\mathcal{M}_{\theta_2}$ works towards finding a stable margin. Once a stable margin is found, $q^{RBL}(\textbf{x})$
The relabeling function (\ref{eqr}) depends on $\theta_2$. However, this dependency can be removed by using the dual representation of the relabeling function. In particular, substituting $\theta_2$ with the optimal dual solution leads to a relabeling function independent of $\theta_2$. We have successfully captured all the spatial details in the reproducing kernel Hilbert space by combining feature and label relationships. This is represented by $| k^T (\textbf{x}) (\textbf{a} \odot \textbf{y})|$ and $|k(\textbf{x}) \odot \textbf{g} - y \mathds{1}_N|$, respectively. Where $k(\textbf{x}) = [k(\textbf{x}_1, \textbf{x}), ..., k(\textbf{x}_N, \textbf{x})]^T$, $\textbf{y} = {(y_1, ..., y_N)}^T$, $\textbf{h} = [h(\textbf{x}_1), ..., h(\textbf{x}_N)]^T$, and $a = (a_1, ..., a_N)^T$ represent Lagrange multipliers, where $a_n > 0$ indicates that $x_n$ is a margin maximizer. So we can now optimize the following:
\begin{equation}
q^{RBL}(\textbf{x}) = (1 - \tau) | k^T (\textbf{x}) (\textbf{a} \odot \textbf{y})| + \tau || k(\textbf{x}) \odot \textbf{h} - y \mathds{1}_N||^{(-1)}
\end{equation}
\begin{figure*}[t]
    \centering
    \subfloat[\centering]{{\includegraphics[width=0.25\textwidth]{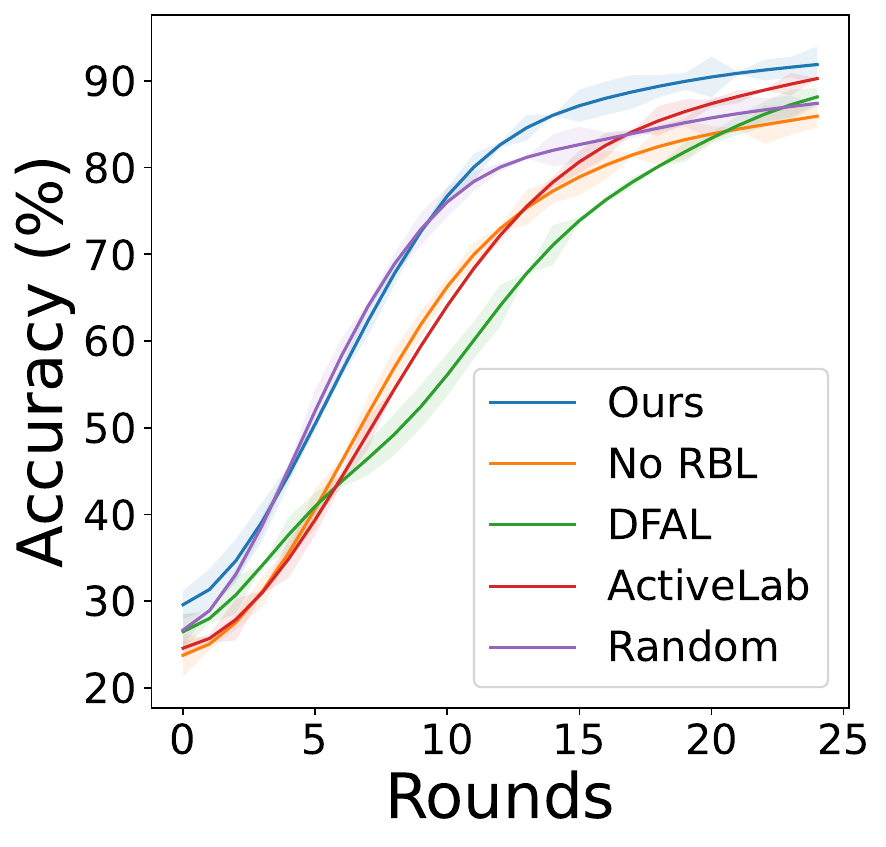} }}%
    \subfloat[\centering]{{\includegraphics[width=0.25\textwidth]{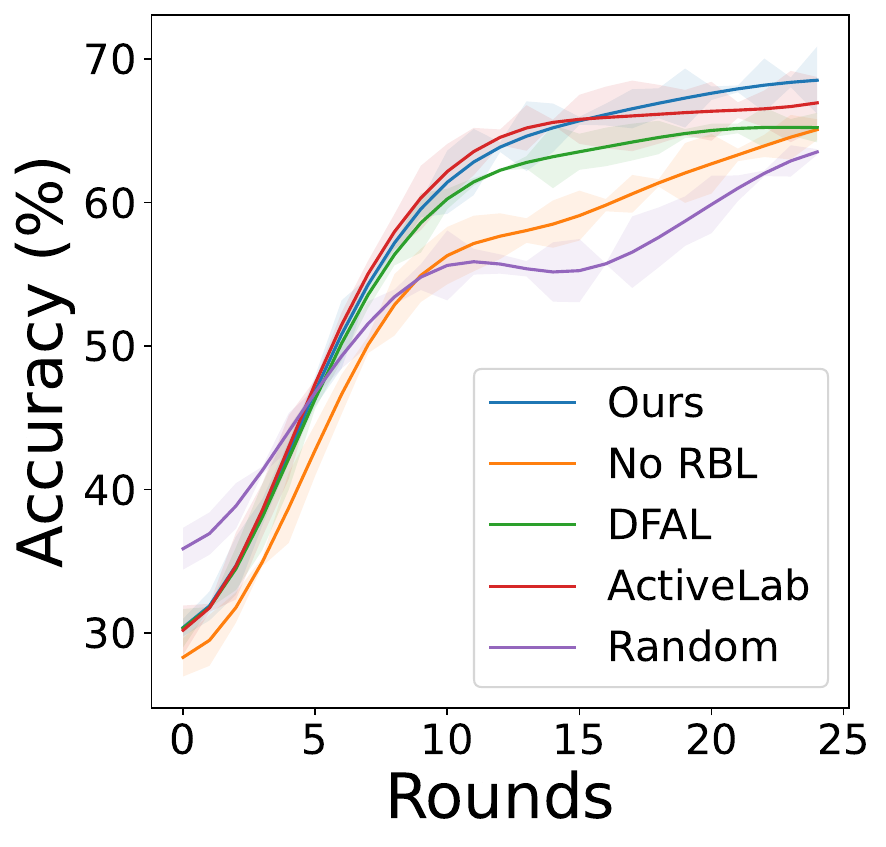} }}%
    \subfloat[\centering]{{\includegraphics[width=0.25\textwidth]{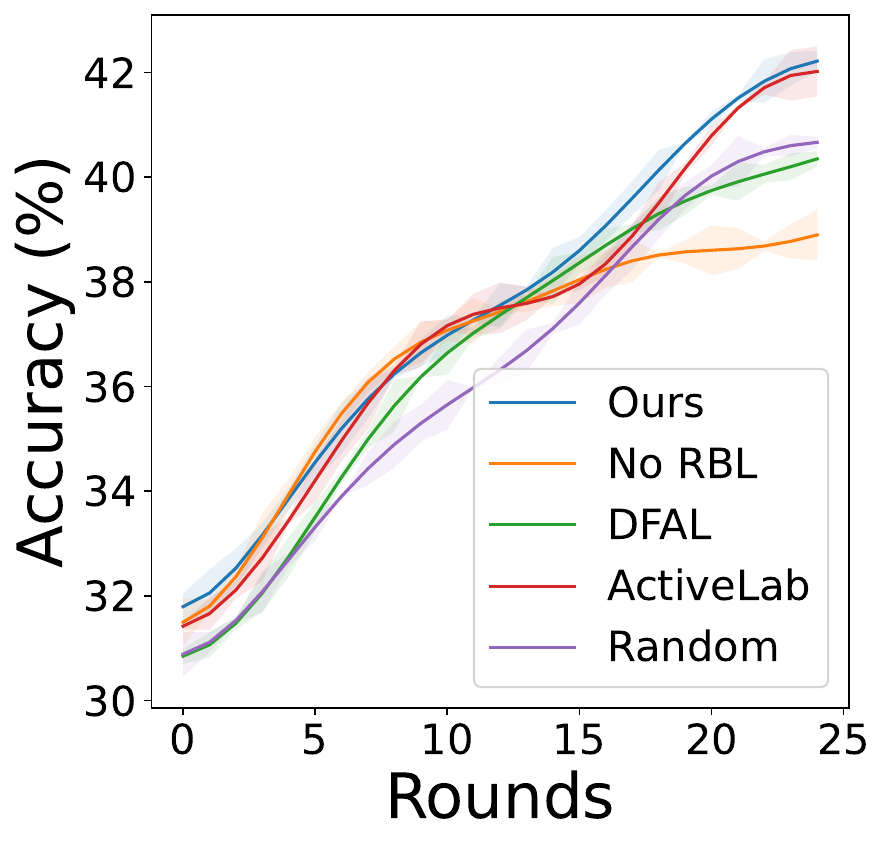} }}%
    \subfloat[\centering]{{\includegraphics[width=0.25\textwidth]{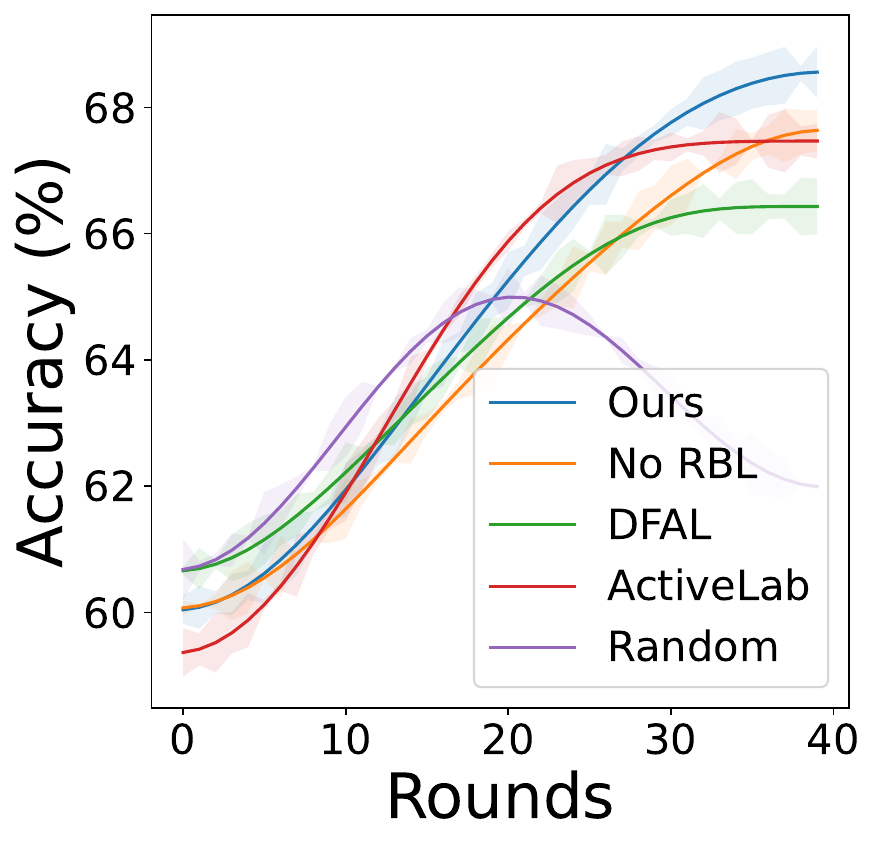} }}%
    \caption{Results of the re-labeling strategy applied to (a) MNIST, (b) FashionMNIST, (c) CIFAR-10, and (d) MedMNIST.}%
    \label{baselines}
\end{figure*}
The samples with the lowest $q^{RBL}$ score will be selected by the model to be re-labeled by the $h_\alpha(.)$ within its annotation budget. However, during our experiment, we noticed that the function $q^{RBL}$ of a clean data instance remains consistently high and stable during the learning process. On the other hand, instances with low $q^{RBL}$ scores usually indicate true noisy data, which will have consistently low scores over multiple learning iterations. This is because the DAL model trained on a few annotated data samples with noise is more likely to have high variance in their predictions for specific samples than a model trained on clean data. This is because noisy annotation makes it difficult for the model to learn the underlying patterns in the data. The increased variance is undesirable and also transfers to our proposed score. To minimize the impact of high variance, we aim to ensure that scores of recently selected samples decrease rapidly. We integrate an exponential moving average into our approach, with a decay factor $\gamma$ that decreases as the DAL learning process progresses.
\begin{equation}
[q^{RBL}(\textbf{x})]^{(r)} = (1 - \gamma) [q^{RBL}(\textbf{x})]^{(r)} 
+ \gamma [q^{RBL}(\textbf{x})]^{(r-1)}
\label{eq8}
\end{equation}
Where $[q^{RBL}(\textbf{x})]^{(r)}$ denotes the score evacuated in the $r_{th}$ round of DAL process.

\section{Experiments}
\begin{table}[t]
\centering
\caption{Detailed settings for different datasets. Frequency denotes the re-labeling frequency within the total annotation budget.}
\begin{tabular}{lccccc}
\hline
\textbf{Dataset} & \textbf{Classes} & \textbf{Train} & \textbf{Test} & \textbf{Initial} & \textbf{Freq.} \\ \hline
\textbf{MNIST} & 10 & 60000 & 10000 & 100 & 2 \\
\textbf{FashionMNIST} & 10 & 60000 & 10000 & 500 & 4 \\
\textbf{CIFAR10} & 10 & 50000 & 10000 & 1000 & 4 \\
\textbf{PathMNIST} & 9 & 89996 & 7180 & 1000 & 4 \\ \hline
\end{tabular}
\label{tab:dttab}
\end{table}
\subsection{Datasets and Setup} \label{DatasetsandSetup}
We evaluated the effectiveness of our approach on four real-world classification datasets: MNIST~\cite{lecun2010mnist}, FashionMNIST~\cite{DBLP:journals/corr/abs-1708-07747} , CIFAR-10 \cite{Krizhevsky09learningmultiple}, and MedMNIST \cite{medmnistv1}. Specifically, we utilized the PathMNIST dataset, which is a subset of MedMNIST2D within the MedMNIST collection. 
Table~\ref{tab:dttab} shows more details about dataset splits. For all four datasets, we divided the original training set into an initial training set and a pool set. Initial samples were randomly selected for training, while the remaining samples formed the pool set for use with the active learning (AL) strategy. We assumed that the initial training set contains noisy labels, which we introduced according to a specified noise ratio. We set the noise rate $\alpha$ to 30\% in our main results and additionally to 10\% and 50\% in the ablation studies. To train our model, we used cross-entropy loss and SGD optimization. The learning rate is set to 0.01, momentum to 0.5, and dropout rate to 10\%. For the re-label strategy, we used $\tau = 0.2$, and with the advancement of DAL rounds, we incrementally increased it to 0.7. The decay factor $\gamma$ is set to 0.2, and the upper limit for re-labeling the same sample multiple times is set to 3. To introduce label noise to the datasets we used the label flip method. We randomly assigned labels from one of the other classes of the dataset. To maintain consistency, we made sure that the new noisy label doesn't match the original class label. Additionally, introducing label noise might imbalance the dataset, but the guaranteed reduction of label noise, as described in Figure \ref{fig2}, prevents such a situation. 

All experiments are performed on workstations with an AMD Ryzen Threadripper PRO 5955WX 16-Core Processor, 2 NVIDIA RTX 6000 GPUs, and 128 GB of RAM. 
\subsection{DAL Acquisition function.} \label{DALAC} The DAL system uses the acquisition function $f(.)$ to determine the next query location. We used uncertainty sampling (\textit{i.e.} smallest margin) with dropout estimation as our acquisition function: \cite{gal2017deep} 
\begin{equation}
   \begin{split}
   f(y = c|\textbf{x}, \mathcal{D}_T) = \int f(y=c|\textbf{x},\theta_1)f(\theta_1|\mathcal{D}_T) d\theta_1 \\
   \approx f(y=c|\textbf{x},\theta_1) \Bar{q}_\theta^*(\theta_1) d\theta_1 \\
   \approx \frac{1}{T} \sum_{t=1}^{T} f(y=c |\textbf{x}, \widehat{\theta_1}_t)
   \end{split}
\end{equation}
with $\widehat{\theta_1}_t$ $\sim$ $\Bar{q}_\theta^*(\theta_1)$, where $\Bar{q}_\theta(\theta_1)$ denotes dropout distribution. We set the dropout rate as 10 in our experiments.
\begin{figure*}[h]
    \centering
    \subfloat[\centering]{{\includegraphics[width=0.265\textwidth]{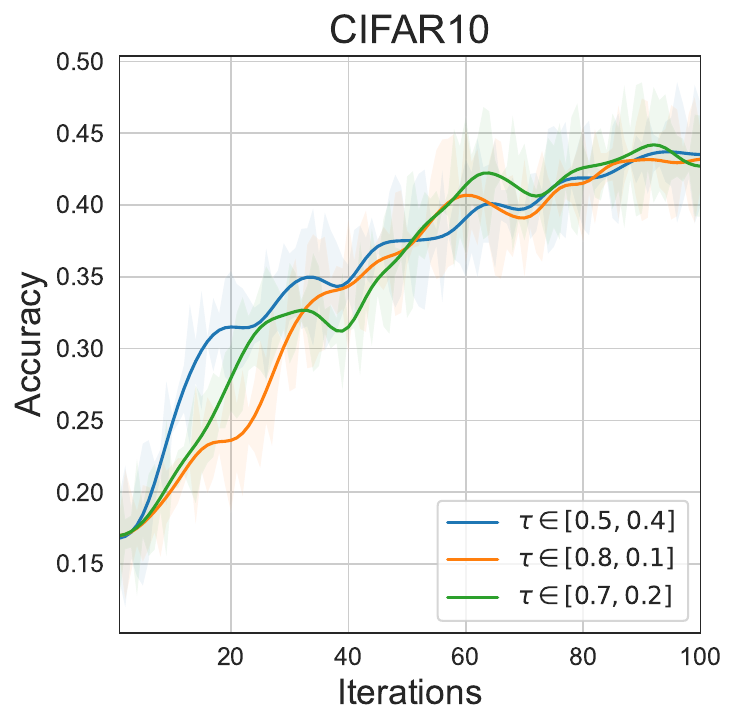} }}
    \subfloat[\centering]{{\includegraphics[width=0.270\textwidth]{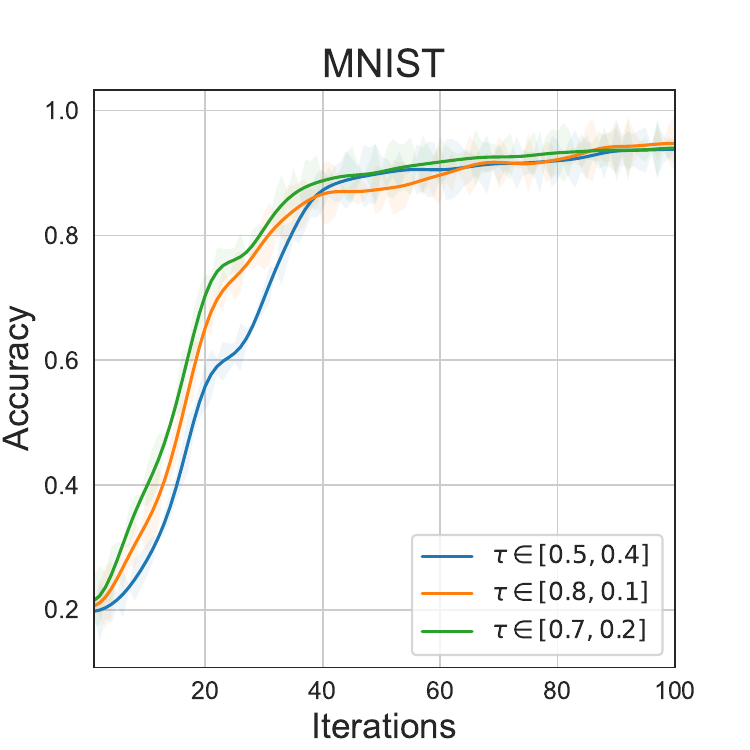} }} 
    \subfloat[\centering]{{\includegraphics[width=0.265\textwidth]{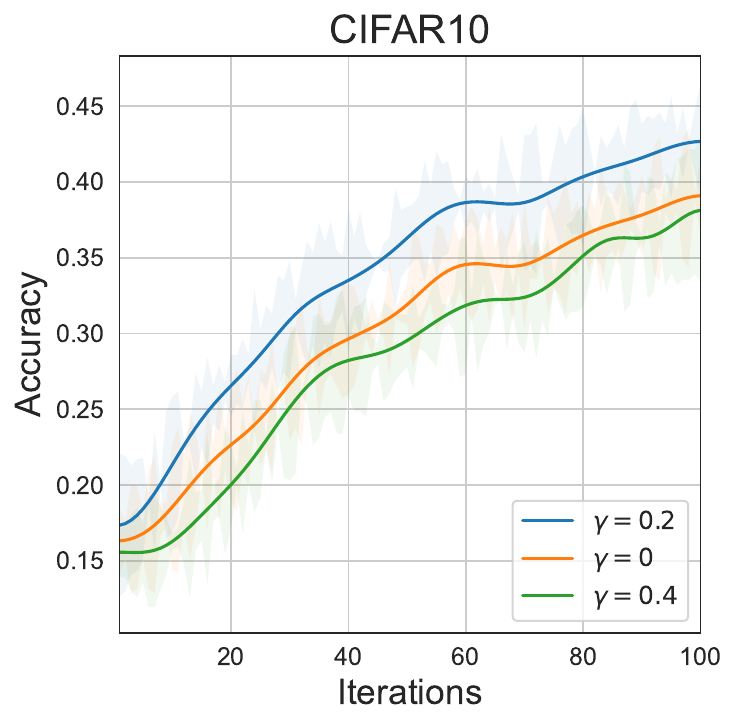} }}
    \\
    \subfloat[\centering]{{\includegraphics[width=0.265\textwidth]{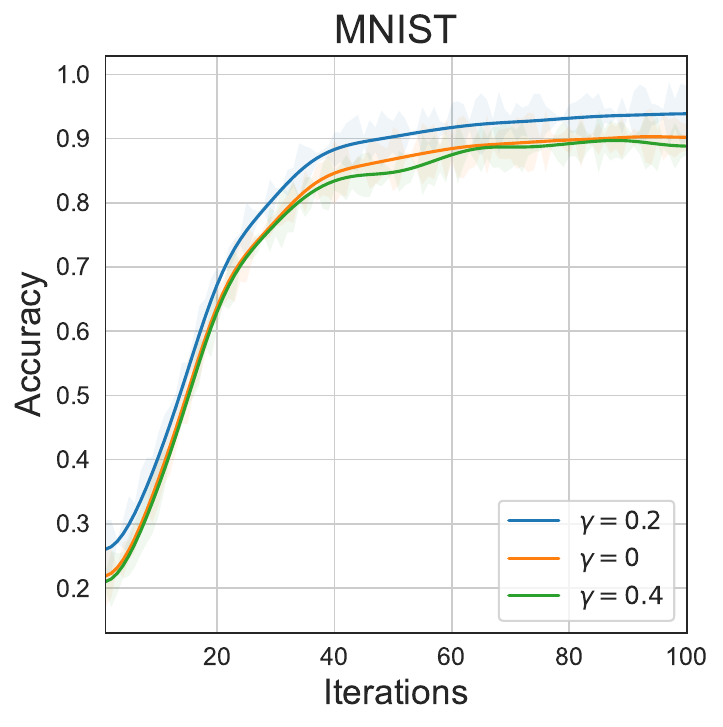} }}
    \subfloat[\centering]{{\includegraphics[width=0.265\textwidth]{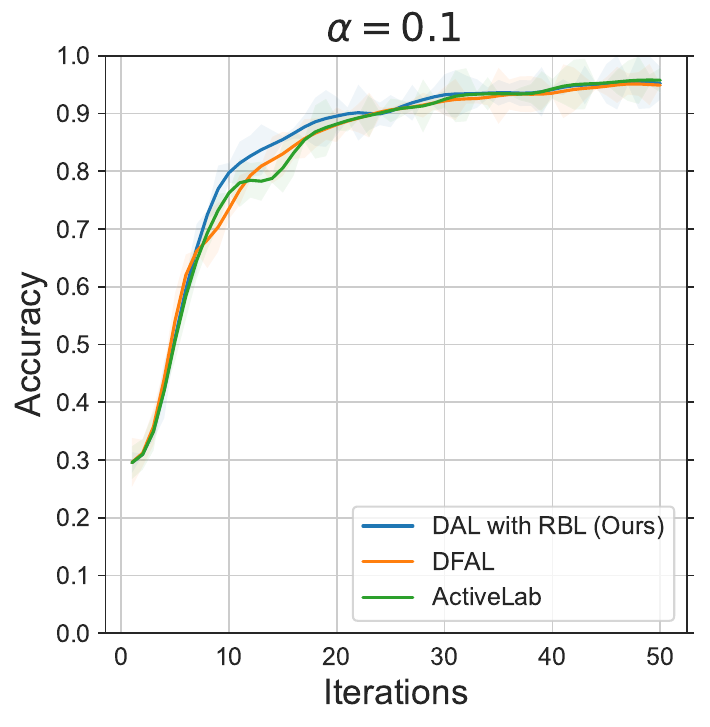} }}%
    \subfloat[\centering]{{\includegraphics[width=0.265\textwidth]{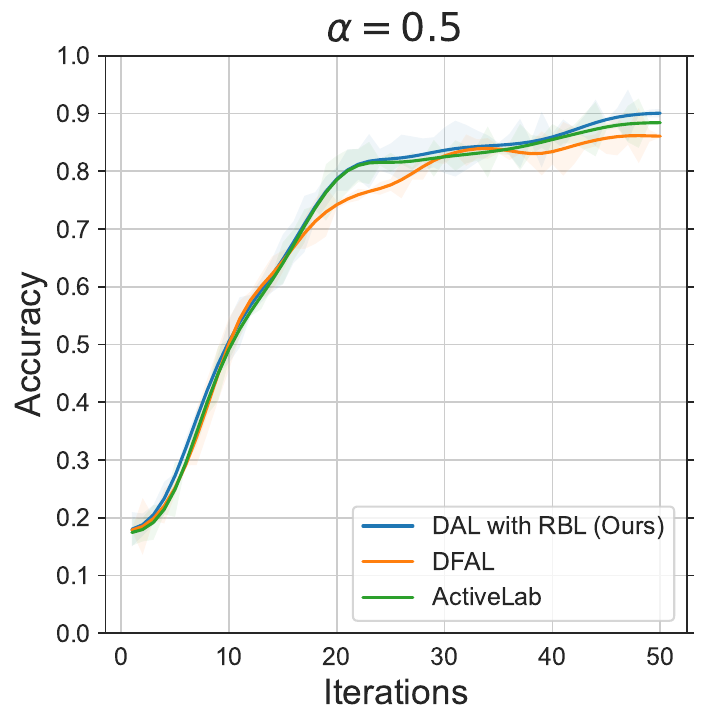} }}%
    \caption{(a-d): Impact of hyperparameters. The impact of hyperparameter $\tau$ on CIFAR10 (a) and MNIST (b) datasets. Impact of the hyperparameter $\gamma$ on CIFAR10 (c) and MNIST (d) datasets. (e-f): Comparison between the performance on a lower noise rate $(\alpha=0.1)$ and higher noise rate $(\alpha=0.5)$ of different baselines on MNIST dataset. All figures display variance information from three runs.}%
    \label{all}
\end{figure*}  
\subsection{Baseline Methods} For comparison, we selected random selection as a baseline to evaluate our proposed strategy. In addition to random selection, we employed the DeepFool Active Learning (DFAL) strategy, which identifies samples with the smallest margins estimated through adversarial attacks \cite{ducoffe2018adversarial}. While developing a novel approach to handling noise in the DAL setting, we found DFAL to be effective in our case, as it leverages a margin-based approach to select valuable samples applicable to various CNN architectures. We also compared our method with ActiveLab \cite{goh2023activelab}, an active learning query strategy designed to select samples for labeling or re-labeling. Although not specifically intended for re-labeling, we applied ActiveLab in this context using a single annotator to ensure a fair comparison. To maintain consistency, we used the same DAL acquisition function—uncertainty sampling (i.e., smallest margin) with dropout estimation \cite{gal2017deep}—across all baselines. Additionally, we included a no-relabeling (No RBL) baseline, which has more annotated samples than other methods since the entire budget is allocated to annotating new samples. To ensure fair comparison, all baselines use the same backbone.
\subsection{Main Results} \label{MainResults}
We report accuracy as the performance measure in line charts for each DAL round, which is conventional in this setting. Figure \ref{baselines}(a)–(d) shows that our strategy consistently outperforms baseline methods across all real-world datasets, each presenting unique challenges. For MNIST, the simplest dataset, our strategy outperforms others from the early stages of the DAL process, with performance steadily improving until convergence. In FashionMNIST, which introduces greater visual diversity, the improvement is more gradual but still surpasses other methods over time. CIFAR-10’s complexity, due to color images and diverse objects, initially challenges our strategy, but after a few DAL rounds, it outperforms the baselines and maintains momentum, unlike the others. A similar trend is observed in MedMNIST, where all methods show comparable performance initially, but as DAL rounds increase, our strategy leverages historical performance through an exponential moving average, effectively addresses margin noise, and dynamically adjusts sample weights, resulting in a clear performance gap. The clear performance gap between our approach and the ``No RBL" baseline emphasizes the critical role of re-labeling in mitigating label noise and enhancing model accuracy. Additionally, although the random strategy performs reasonably well in the early stages, particularly in MedMNIST, it fails to maintain momentum over time, underscoring the importance of strategic sample selection for re-labeling in deep active learning settings. 

\begin{table*}[h]
\centering
\caption{Performance comparison of DAL with re-labeling and random re-labeling at different learning stages with a fixed noise rate ($\alpha = 0.30$) for all datasets. The hit rate and label correction (LC) rate take into account all noisy data samples present up to a specific round of DAL.}
\resizebox{.99\textwidth}{!}{
\begin{tabular}{l|cccccccccccc}
\hline
                       & \multicolumn{6}{c|}{\textbf{DAL-RBL}}                                                                                                                                              & \multicolumn{6}{c}{\textbf{Random}}                                                                                                                           \\ \hline
\textbf{Dataset}       & \multicolumn{2}{c|}{\textbf{Hit Rate}}                    & \multicolumn{2}{c|}{\textbf{LC Rate}}                     & \multicolumn{2}{c|}{\textbf{Accuracy}}                    & \multicolumn{2}{c|}{\textbf{Hit Rate}}                    & \multicolumn{2}{c|}{\textbf{LC Rate}}                     & \multicolumn{2}{c}{\textbf{Accuracy}} \\ \cline{2-13} 
                       & \multicolumn{1}{c|}{Initial} & \multicolumn{1}{c|}{Later} & \multicolumn{1}{c|}{Initial} & \multicolumn{1}{c|}{Later} & \multicolumn{1}{c|}{Initial} & \multicolumn{1}{c|}{Later} & \multicolumn{1}{c|}{Initial} & \multicolumn{1}{c|}{Later} & \multicolumn{1}{c|}{Initial} & \multicolumn{1}{c|}{Later} & \multicolumn{1}{c|}{Initial} & Later  \\ \hline
\textbf{MNIST}         & 0.25                        & 0.43                      & 0.25                       & 0.39                      & \textbf{0.82}                & \textbf{0.95}             & 0.57                        & 0.16                      & 0.14                       & 0.11                      & 0.80                       & 0.88  \\
\textbf{Fashion MNIST} & 0.14                        & 0.23                      & 0.14                       & 0.20                      & \textbf{0.62}               & \textbf{0.74}             & 0.64                        & 0.15                      & 0.16                       & 0.10                      & 0.58                        & 0.69 \\
\textbf{CIFAR10}       & 0.17                        & 0.21                      & 0.19                       & 0.17                      & \textbf{0.37}               & \textbf{0.47}             & 0.58                        & 0.17                      & 0.15                       & 0.12                      & 0.33                        & 0.43   \\ \hline
\end{tabular}
}
\label{table1}
\vskip -0.10in
\end{table*}
\subsection{Evaluation with Different Performance Measures}  
We evaluate our re-labeling strategy using \textbf{hit rate} and \textbf{label correction (LC) rate}, which quantify the proportion of correctly identified and corrected noisy samples, respectively. As Table~\ref{table1} shows, in the early stage of DAL, our method identifies fewer noisy samples than random selection yet achieves higher accuracy. This reflects its ability to prioritize correcting the most detrimental noise while tolerating outliers, so that even with a lower hit rate, label corrections are more effective. These findings highlight the efficacy of our targeted re-labeling approach, demonstrating that strategic noise correction can substantially enhance model performance, particularly in the initial stages of DAL.

Table \ref{table2} presents additional baseline comparisons on the MNIST dataset. The results are consistent with the observations in Table \ref{table1}. Although DFAL and ActiveLab show relatively strong initial performance, their effectiveness decreases over time, likely due to the absence of an ordered sample-selection strategy. In contrast, our method continues to improve throughout the learning process and ultimately surpasses all baselines in the later stage, demonstrating superior robustness and long-term effectiveness in DAL settings.
\subsection{Analysis and Ablation Studies}
\subsubsection{Impact of hyperparameters.} We conducted an ablation study on the newly introduced hyperparameters. The hyperparameter $\tau$ balances the $q^{DEC}(\textbf{x})$ and $q^{INC}(\textbf{x})$ functions. As both $q^{DEC}(\textbf{x})$ and $q^{INC}(\textbf{x})$ are normalized, the range of $\tau$ is within [0,1]. We experimented with the range of $[0.1, 0.8]$, $[0.2, 0.7]$ and $[0.5, 0.6]$. Our results show that $\tau$ is robust in the range of $[0.1, 0.8]$ and $[0.2, 0.7]$, and the lowest accuracy was achieved in the range of $[0.5, 0.6]$. This finding aligns with the analysis in Section \ref{MainResults}. The experimental results on two datasets are displayed in Figure \ref{all}(a-b). We also introduced another hyperparameter, $\gamma$, which is the exponential moving average. The hyperparameter $\gamma$ aims to reduce learning variance by minimizing the score of recently selected samples. Figure \ref{all}(c-d) displays the performance comparison for different $\gamma$ values on two datasets. We experimented with three different values of $\gamma$ and found that a higher value of $\gamma$ leads to lower performance, consistent with our design of equation \ref{eq8}. Overall figure Figure \ref{all}(a-d) demonstrates the robustness of our strategy, indicating minimal sensitivity to hyperparameters.

\subsubsection{Impact of different noise rates.}
We compared the performance of our method and other baselines at two different noise rates ($\alpha$). We observed that at a lower annotator noise rate ($\alpha = 0.1$), our method outperformed the other baselines, although the difference was minimal (see Figure \ref{all}(e)). This is because with a low noise rate, there is limited opportunity for improvement, and the combination of the DAL model and query strategy helps to narrow the gap. As a result, all three re-label baselines perform similarly. The performance gap and gain are clearly visible for a higher noise rate ($\alpha = 0.5$). Our method consistently outperformed the other two baselines and yielded a stable active learning curve (see Figure \ref{all}(f)). The overall Figure \ref{all}(e-f) demonstrates the impact of noise in DAL and the effectiveness of our method in mitigating the impact at lower and higher label noise rates.

\subsubsection{Impact of DEC and INC} 
\begin{figure}[h]
    \centering
    \subfloat[\centering]{{\includegraphics[width=0.24\textwidth]{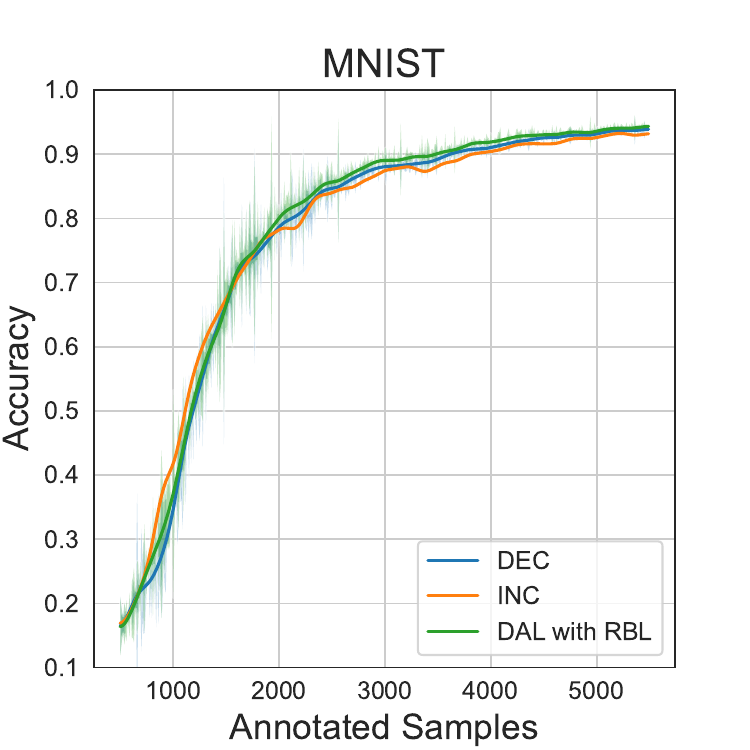} }}%
    \subfloat[\centering]{{\includegraphics[width=0.24\textwidth]{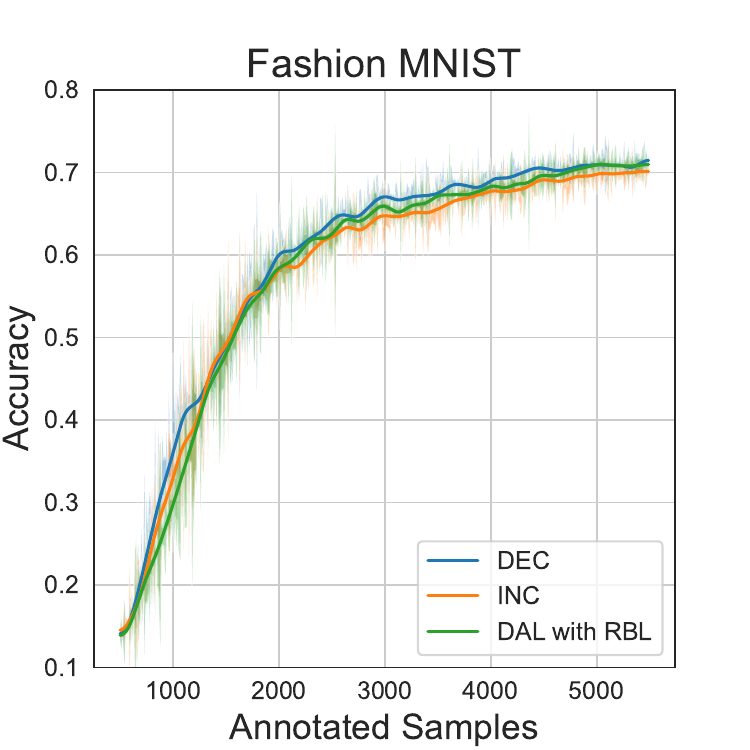} }}%
    \caption{Comparison between the performance of two different components of our strategy (\textit{i.e.} DEC, INC) and the combined proposed strategy (\textit{i.e.} DAL with RBL).}%
    \label{fg6}
    \vskip -0.15in
\end{figure}

We analyzed our strategy by comparing the performance of the two main components, DEC and INC, at different stages of DAL. Figure \ref{fg6} illustrates the comparison results on two real-world datasets. In Figure \ref{fg6}(a)-(b), we observed that during the initial rounds of DAL, INC outperformed DEC. This can be attributed to the fact that the deep classifier was still in the early stages of learning and adjusting the margin to achieve better separation of classes. In contrast, INC focused on selecting samples that were further from the decision boundary and inconsistent. Although this approach performed well in the early stages, it did not contribute to better separation of the margin in the long run. Ultimately, DEC proved to be more effective than INC as the margin became more stable after a few DAL rounds. Our proposed strategy incorporates the benefits of both DEC and INC, leading to overall better performance and faster model convergence. Our sample selection order for re-labeling is further validated by this result.
\begin{table}[t]
\centering
\caption{Performance Comparison with additional baselines using MNIST dataset.}
\resizebox{1.00\columnwidth}{!}{
\begin{tabular}{l|cccccc}
\hline
\textbf{Strategy}      & \multicolumn{2}{c|}{\textbf{Hit Rate}}                    & \multicolumn{2}{c|}{\textbf{LC Rate}}                     & \multicolumn{2}{c}{\textbf{Accuracy}} \\ \cline{2-7} 
                       & \multicolumn{1}{c|}{Initial} & \multicolumn{1}{c|}{Later} & \multicolumn{1}{c|}{Initial} & \multicolumn{1}{c|}{Later} & \multicolumn{1}{c|}{Initial}  & Later \\ \hline
\textbf{Random}        & 0.57                        & 0.16                      & 0.14                        & 0.11                      & 0.80                         & 0.88 \\
\textbf{DFAL}      & 0.24                        & 0.20                      & 0.24                        & 0.15                      & 0.80                         & 0.89 \\ 
\textbf{ActiveLab}      & 0.25                        & 0.22                      & 0.25                        & 0.28                      & 0.78                         & 0.91 \\
\textbf{Ours} & 0.25                        & 0.43                      & 0.25                        & 0.39                      & 0.82                         & 0.95 \\
\hline
\end{tabular}
}
\label{table2}
\end{table}
\begin{table}[h]
\centering
\large
\caption{Performance Comparison Across Rounds on PairFlip Label Noise.}
\resizebox{\columnwidth}{!}{
\begin{tabular}{lccccc}
\toprule
\textbf{Method} & \textbf{Round 1} & \textbf{Round 15} & \textbf{Round 25} & \textbf{Round 35} & \textbf{Round 50} \\
\midrule
\textbf{No Relabeling} & 16.78 & 36.49 & 47.87 & 68.45 & 70.14 \\
\textbf{Random} & 16.78 & 44.22 & 58.36 & 77.33 & 74.21 \\
\textbf{DFAL} & 16.78 & 39.41 & 62.79 & 83.42 & 80.90 \\
\textbf{ActiveLab} & 16.78 & 33.90 & 72.77 & 83.86 & 83.67 \\
\textbf{Ours} & 16.78 & 47.92 & 68.17 & 81.89 & 84.86 \\
\bottomrule
\end{tabular}
}
\label{fliptable}
\vskip -0.20in
\end{table}
\subsection{Robustness to Non-Uniform Annotation Noise} \label{flipt}
We further evaluate our method under a more challenging and realistic noise setting, where annotators are prone to confuse visually similar classes. For example, the digit ``7'' is frequently mislabeled as ``1'' due to their structural similarity. To simulate this, we adopt a class-dependent pair-flip scheme: $0 \rightarrow 8$, $1 \rightarrow 7$, $2 \rightarrow 3$, $3 \rightarrow 2$, $4 \rightarrow 9$, $5 \rightarrow 6$, $6 \rightarrow 5$, $7 \rightarrow 1$, $8 \rightarrow 0$, $9 \rightarrow 4$. This setup introduces structured, non-uniform errors that more closely mimic real human annotation behavior compared to purely random label noise. 

Table~\ref{fliptable} compares different strategies under this setting on MNIST. The ``No Relabeling'' baseline allocates the full budget to new samples, leading to gradual improvement but leaving mislabeled points uncorrected. Methods such as DFAL and ActiveLab, which explicitly target re-labeling, show gains but plateau due to their single-objective design that limits sustained noise correction. In contrast, our approach dynamically balances exploration (new samples) with exploitation (targeted re-labeling), yielding consistent improvements across training stages. Specifically, it corrects early-stage decision-boundary noise and continues to mitigate inconsistent label noise in later stages. 

These results demonstrate that our method is not only robust to uniform random noise but also effective under structured, class-dependent noise, a setting that better reflects real-world annotation challenges. By addressing both boundary-critical errors and broader inconsistencies, our framework achieves more stable performance compared to the baselines.

\section{Discussions and Limitations} Our relabeling strategy is systematically designed for a restricted setting with a limited annotation budget. It aims to detect and select the most informative samples with noisy labels while sticking to budget constraints. In such a scenario, both the majority voting and student-teacher methods impose additional burdens on the budget. Moreover, the majority voting approach heavily relies on expert weighting, typically assigning equal weight to each expert. However, in real-world settings, maintaining an annotation budget is crucial for DAL, and accurately weighing experts according to their expertise for a downstream task can be challenging. Although our strategy is designed to utilize a single annotator, it can also accommodate multiple annotators for majority voting, offering additional advantages for our approach.

\section{Conclusion}
We proposed a framework for deep active learning, which
includes a unique strategy called deep active re-labeling. Before
developing the framework, we conducted both theoretical
and numerical analyses to ensure its effectiveness. Our analysis
showed that our approach can reduce errors at a guaranteed
rate. To achieve this, we combined MMC and deep learning
models in our re-labeling process. We have incorporated an
exponential moving average into the strategy to reduce its
susceptibility to high levels of noise variance. By focusing on
selecting the most informative and wrongly annotated samples,
our approach outperformed all other baseline methods. We
also conducted extensive experiments and studies on our
framework, which supported our analysis. Our approach is
practical for real-world applications as it utilizes the same
Oracle annotator for re-labeling purposes. Moving forward,
we plan to investigate more sophisticated deep learning model
structures, such as attention mechanisms in transformers,
within our framework.
\bibliographystyle{IEEEtran}
\bibliography{mybibfile}

\end{document}